\documentclass[letterpaper, 10 pt, conference]{ieeeconf}  

\IEEEoverridecommandlockouts                              

\usepackage{cite} 
\usepackage{graphicx}
\usepackage{array}      
\usepackage{booktabs}
\usepackage{graphicx} 
\usepackage{tabularx}
\usepackage{booktabs}
\usepackage{multirow}
\usepackage{amsmath}   
\usepackage{amssymb}   

\title{\LARGE \bf
Defending Deepfake via Texture Feature Perturbation}

\author{Xiao Zhang$^{1}$, Changfang Chen$^{2}$ and Tianyi Wang$^{3,*}$%
\thanks{This study is grateful to the Key R\&D Program of Shandong Province (Major Scientific and Technological Innovation Project): (Grant No.2023CXGC010113); The Taishan Scholars Program: (Grant NO.tspd20240814); Qilu University of Technology (Shandong Academy of Sciences) Project (Grant No.2024ZDZX11); Key Laboratory of Computing Power Network and Information Security, Ministry of Education, Qilu University of Technology
(Shandong Academy of Sciences). }%
\thanks{*Corresponding author: Tianyi Wang.}
\thanks{$^{1}$Xiao Zhang is with the Key Laboratory of Computing Power Network and Information Security, Ministry of Education, Qilu University of Technology (Shandong Academy of Sciences), Jinan, China
        {\tt\small 10431230097@stu.qlu.edu.cn}}%
\thanks{$^{2}$Changfang Chen is with the Shandong Artificial Intelligence Institute, Qilu University of Technology (Shandong Academy of Sciences), Jinan, China
        {\tt\small chenbhlt012@qlu.edu.cn}}%
\thanks{$^{3}$Tianyi Wang is with the School of Computing, National University of Singapore
        {\tt\small wangty@nus.edu.sg}}%
}

\begin{document}
\maketitle
\thispagestyle{empty}
\pagestyle{empty}

\begin{abstract}

The rapid development of Deepfake technology poses severe challenges to social trust and information security. While most existing detection methods primarily rely on passive analyses, due to unresolvable high-quality Deepfake contents, proactive defense has recently emerged by inserting invisible signals in advance of image editing. In this paper, we introduce a proactive Deepfake detection approach based on facial texture features. Since human eyes are more sensitive to perturbations in smooth regions, we invisibly insert perturbations within texture regions that have low perceptual saliency, applying localized perturbations to key texture regions while minimizing unwanted noise in non-textured areas. Our texture-guided perturbation framework first extracts preliminary texture features via Local Binary Patterns (LBP), and then introduces a dual-model attention strategy to generate and optimize texture perturbations. Experiments on CelebA-HQ and LFW datasets demonstrate the promising performance of our method in distorting Deepfake generation and producing obvious visual defects under multiple attack models, providing an efficient and scalable solution for proactive Deepfake detection.

\end{abstract}

\begin{figure*}[t] 
\vspace*{10pt} 
\centering
\includegraphics[width=0.9\textwidth]{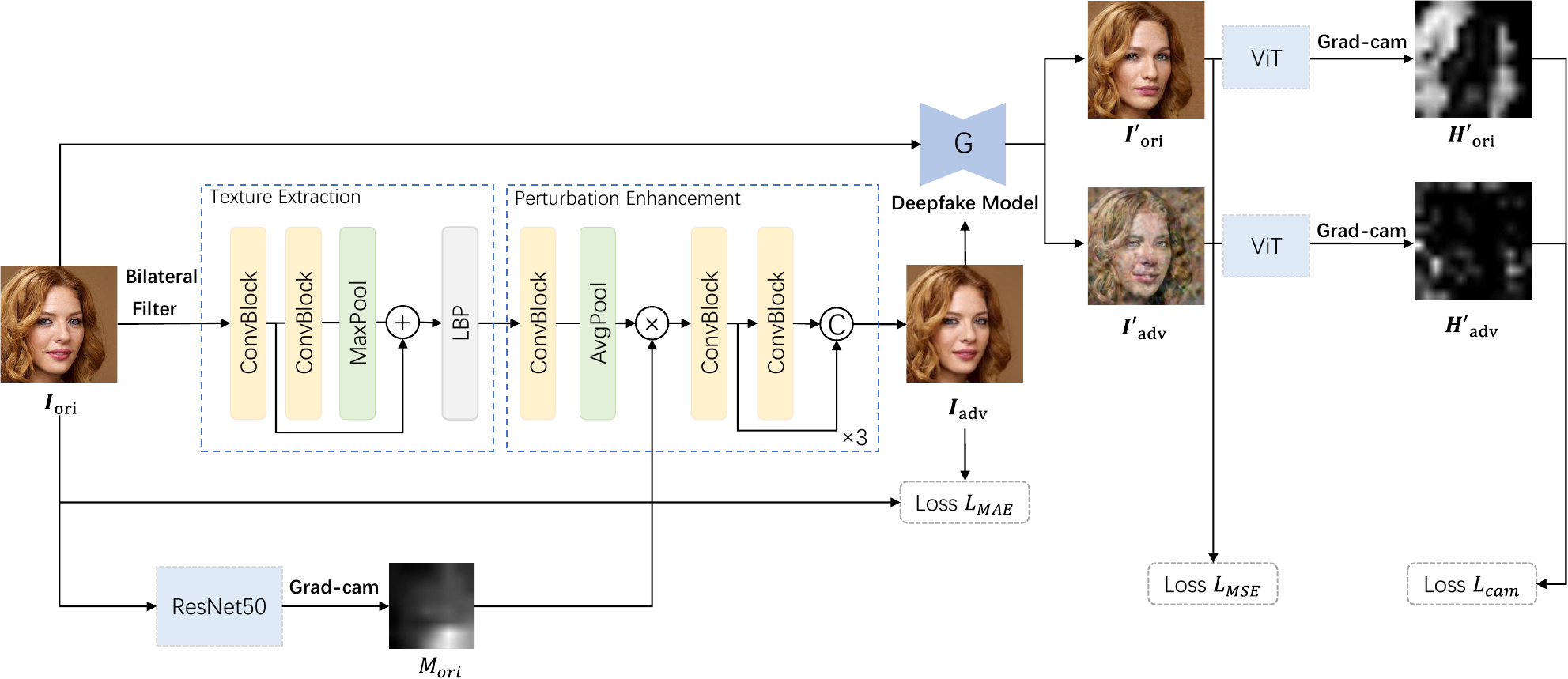}
\caption{Demonstration of the proposed framework. For the input image $I_\textrm{ori}$, the texture extraction module extracts its initial texture information and inputs it into the ResNet50 network, where Grad-CAM is applied to generate the corresponding feature attention map $M_\textrm{ori}$, the obtained $M_\textrm{ori}$ and the initial texture information are enhanced in the perturbation enhancement module. The generated perturbed image exhibits obvious visual defects after passing through the Deepfake model.} 
\label{fig1}
\end{figure*}
\section{INTRODUCTION}

With the continuous improvement of generative algorithms, modern Deepfake technologies~\cite{chen2020simswap,ren2023reinforced,zhou2023uniface} can produce synthetic images that are indistinguishable to human eyes, with highly realistic facial features, voice timbre, and body movements. However, malicious usages of Deepfake have posed serious and urgent threats to the authenticity of media information~\cite{juefei2022countering}. 

Countermoves for detection has been proposed since the first occurrence of Deepfake contents~\cite{wang2024deepfake}. They generally fall into two categories: passive detection and proactive defense. Traditional passive detection~\cite{guo2021fake,wang2023deep,wang2023noise,yang2022multi} primarily focuses on identifying whether content is generated by Deepfake techniques. In contrast, proactive defense emphasizes preventing Deepfake models from generating convincing forged content by embedding imperceptible defense information or changing data features before the forged content is generated.

Existing proactive defense methods are primarily divided into watermark-based~\cite{beuve2023waterlo,wang2021faketagger,IDPMark2024Wang,wang2024lampmark,wang2025fractalforensics} and perturbation-based~\cite{aneja2022tafim,he2022defeating,huang2022cmua,NullSwap2025Wang} approaches. Watermarking embeds identifiable markers for copyright protection and traceability, however, the embedded information can be fragile under white-box attacks, making it susceptible to removal 
or overwriting by adversaries. In contrast, perturbation-based methods directly distort the generation results of Deepfake models by introducing subtle modifications that are imperceptible to the human eye.
Existing perturbation defense methods that perform perturbation embedding in the spatial domain typically adopt a uniform noise injection strategy for the image. However, this indiscriminate perturbation pattern has two key flaws. First, since the human visual system is more sensitive to noise in flat areas than in edges or textured regions~\cite{luo2022frequency}, the uniform injection strategy will cause noise to produce obvious visual artifacts in smooth areas, seriously affecting image quality. On the other hand, this method may result in insufficient perturbation intensity in key facial texture areas, making it difficult to effectively suppress the generative model's ability to imitate these areas.

To address the aforementioned issues, this study proposes a perturbation framework guided by texture features. For an input image, we first preprocess it through bilateral filtering, extract basic texture features through local binary patterns in the texture extraction module, and generate attention feature maps using the gradient-weighted class activation mapping (Grad-CAM) technique. These two feature maps are then fused in the perturbation enhancement module, and a perturbed image is generated through multi-layer deformable convolution operations. Thereafter, the forged image and its corresponding attention feature map are derived. To optimize the perturbation effect, we construct a multi-objective loss function by comparing the feature differences between the original and the perturbed output. This effectively interferes with the forged generation process while ensuring visual quality. The contributions of this work are threefold:

\begin{itemize}
\item[$\bullet$] We propose a perturbation framework based on facial texture features. By accurately identifying and perturbing the texture features of key facial areas, the framework effectively destroys the imitation and generation of key textures in the Deepfake generation process.
\item[$\bullet$] We design a dual-model attention strategy to guide the optimization of perturbations. This method integrates initial texture features with attention maps generated through local feature parsing for regional enhancement while leveraging global semantic modeling to optimize perturbation directions. By optimizing perturbations across multiple spatial scales, we achieve a balance between defense effectiveness and visual quality. 
\item[$\bullet$] Extensive experiments demonstrate that the framework can effectively resist Deepfake manipulation models and produce perturbed results with significant visual defects.
\end{itemize}

\begin{figure*}[htbp]  
\centering
\includegraphics[width=\textwidth]{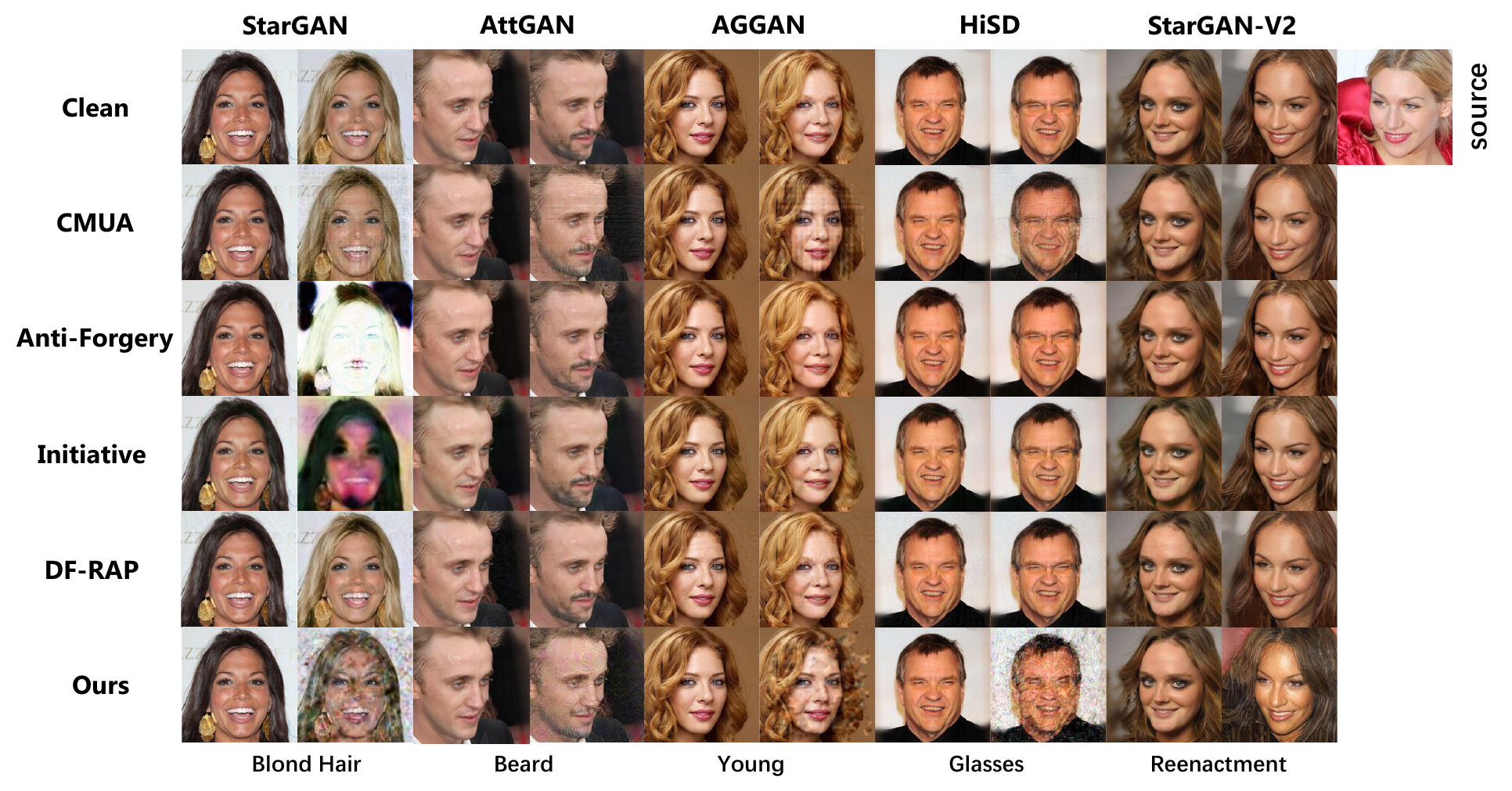}
\caption{Visual effects of Deepfake attribute editing and reenactment image manipulations on the different defense methods. The source image in the last column is dedicated to providing expression features for StarGAN-V2 when reenacting. In the remaining columns, all odd-numbered columns display perturbation images while even-numbered columns show the corresponding Deepfake results.} 
\label{fig2}
\end{figure*}
\section{RELATED WORK}
\subsection{Deepfake Generation}

Deepfake attribute editing modifies specific facial attributes in images or videos while preserving the person's identity, achieving high-quality and natural results. StarGAN~\cite{choi2018stargan} employs a unified multi-domain architecture for cross-attribute image conversion without separate domain-pair training. AttentionGAN~\cite{tang2019attention} enhances local detail editing by using spatial attention maps to isolate target regions, while AttGAN~\cite{he2019attgan} disentangles attribute and identity features, enforcing target attributes via classification loss. HiSD~\cite{li2021image} achieves fine-grained control through hierarchical style disentanglement, separating semantically relevant and irrelevant layers.

For expression reenactment, StarGAN-V2~\cite{choi2020stargan} eliminates the need for explicit attribute labels by transferring styles from a source image to a target reference image, enabled by its mapping network and style encoder. HyperReenact~\cite{bounareli2023hyperreenact} further refines this by leveraging StyleGAN2's~\cite{karras2020analyzing} photorealistic generation, using a hypernetwork to manipulate latent spaces for precise identity preservation and pose retargeting, avoiding artifacts from external tools.

\subsection{Deepfake Defense}
Deepfake generation defense strategies can be divided into passive detection and proactive defense. Passive detection~\cite{li2021frequency,li2020face,shiohara2022detecting,xu2024fakeshield,zhao2021multi} typically determines the authenticity of an image by identifying anomalies in the spatial or frequency domain. To overcome the shortcomings of passive detection, perturbation techniques use specific technical means in the Deepfake generation process. These methods intentionally introduce distortions and forgeries, effectively reducing the quality of synthetic images.

CMUA-Watermark~\cite{huang2022cmua} aims to generate universal adversarial perturbations that are effective across multiple models without relying on any specific Deepfake model. Anti-Forgery~\cite{wang2022anti} generates adversarial perception-sensitive perturbations in the Lab color space, which exhibit strong robustness against various input transformation operations. TAFIM~\cite{aneja2022tafim} optimizes a global perturbation pattern and a conditional generative model, where an attention network and a fusion network combine multiple model-specific perturbations to generate perturbations tailored to specific images. Meanwhile, TCA-GAN~\cite{dong2023restricted} introduces a transferable adversarial attack that effectively disrupts Deepfake models in black-box settings. Guan et al.~\cite{guan2022defending} proposed an ensemble defense strategy called Hard Model Mining (HMM). This strategy does not optimize for good average performance during the perturbation optimization process. Instead, it improves the perturbation's ability to disrupt the most resilient models at each iteration step, thereby enabling it to disrupt all Deepfake models. To further enhance the generalization and robustness of perturbations, ComGAN~\cite{qu2024df} was proposed to learn the shared characteristics of compression across different platforms to generate robust adversarial perturbations.

\section{METHODOLOGY}

\subsection{Method Overview}
To satisfy the dual objectives of adversarial perturbation and visual quality, in this paper, we propose a perturbation method based on facial texture features. Unlike traditional uniform perturbation strategies, we embed perturbations within the image's textured regions. The naturally complex structures and high-frequency details in these regions provide an ideal context for concealing subtle perturbations or modifications. Furthermore, perturbations in texture regions disrupt the learning process of generative models on these critical features, causing distortions of the generated images and thereby degrading the overall generation quality.

Fig.~\ref{fig1} shows the design of the model. 
The input raw image $I_\textrm{ori}$ is processed through two parallel branches. In the first branch, bilateral filtering~\cite{tomasi1998bilateral} is applied to eliminate high-frequency noise interference, followed by a texture extraction module to obtain deep texture features. In the second branch, $I_\textrm{ori}$ is fed directly into a ResNet50 network~\cite{he2016deep}, where the Grad-CAM algorithm~\cite{selvaraju2017grad} is employed to generate the feature attention distribution $M_\textrm{ori}$. Subsequently, a perturbation enhancement module is designed to spatially enhance the initial texture features guided by the feature attention mechanism. This process ultimately produces visually imperceptible adversarial perturbations that can produce noticeable distortions after being tampered with by the Deepfake model.

\subsection{Texture Extraction}
We use Bilateral Filter combined with Local Binary Pattern (LBP)~\cite{ojala1994performance} for texture extraction. The core steps include grayscale conversion, bilateral filter preprocessing, and LBP texture quantization.

\noindent\textbf{Grayscale Conversion}. Texture feature analysis is typically based on luminance information. Therefore, we convert the original image $I_\textrm{ori}\in\mathbb{R}^{H\times W\times3}$ to grayscale to eliminate color interference, using weighted average to compute the luminance channel:
\begin{equation}
I_{\mathrm{gray}}(x,y) = 0.299 R(x,y) + 0.587 G(x,y) + 0.116 B(x,y),
\end{equation}
where $(x, y)$ represents the pixel coordinates, and the weight coefficients comply with the ITU-R BT.709 standard~\cite{bt2002parameter}, ensuring that the converted grayscale image $I_{\textrm{gray}}$ aligns with the human visual brightness perception characteristics.

\noindent\textbf{Bilateral Filter}. We apply a bilateral filter to the grayscale image in order to smooth noise while preserving texture edges. The filtered image $I_\textrm{filter}$ suppresses noise in flat regions while enhancing high-frequency texture details. The bilateral filtering formula is denoted as follows: 
\begin{equation}
I_{\mathrm{filter}}(x,y) = \frac{1}{W_{\mathrm{BF}}} \sum_{i,j \in \Omega} 
G_{\sigma_d}(i,j) \cdot G_{\sigma_r}(\Delta I) \cdot I_{\mathrm{gray}}(x + i, y + j),
\label{eq:bilateral_filter}
\end{equation}

\begin{equation}
\Delta I = \bigl| I_{\mathrm{gray}}(x,y) - I_{\mathrm{gray}}(x+i,y+j) \bigr|,
\end{equation}
where $\Omega$ represents the neighborhood window range, $\sigma_d$ is the standard deviation of the spatial Gaussian kernel, and $\sigma_r$ is the standard deviation of the range Gaussian kernel.

\noindent\textbf{LBP Texture Quantization}. The core of this process is to apply two ConvBlocks to the filtered image $I_\textrm{filter}$ to extract multi-level structural features. After compressing the spatial information using max pooling, the LBP method is employed, with the formula as follows:
\begin{equation}
LBP_{P,R}(x_c, y_c) = \sum_{p=0}^{P-1} s(g_p - g_c) 2^p,
\end{equation}
where $P$ denotes the number of sampling points in the circular neighborhood, $R$ denotes the radius of the neighborhood, $g_c$ is the grayscale value of the central pixel, $g_p$ is the grayscale value of the 
p-th sampling point, and $x_c$ and $y_c$ represent the coordinates of the central pixel in the image. When traversing the entire image, each pixel becomes the central pixel $(x_c,y_c)$ once, and its LBP value is calculated using this formula. $s(x)$ is the sign function, defined as follows: 
\begin{equation}
 s(x) = 
\begin{cases} 
1 & \text{if } x \geq 0, \\
0 & \text{otherwise}.
\end{cases}
\end{equation}

\subsection{Perturbation Enhancement}
In order to ensure that the perturbation focuses on the texture regions while reducing the unwanted noise in the smooth area, we introduced an attention guidance mechanism based on Grad-CAM, and enhanced the initial texture area through a ConvBlock. A single ConvBlock consists of a convolutional neural network (CNN) layer, a batch normalization layer, and a ReLU activation function. The feature map output by the convolution block is spatially compressed using average pooling. At the same time, we use the pre-trained classification model to forward propagate the input image $I_\textrm{ori}$, generate the feature attention map $M_\textrm{ori}$ through reverse gradient calculation, and perform channel-by-channel weighted fusion with the multi-scale texture features after spatial compression to ensure that the perturbation is aligned with the model's attention area. This fusion strategy ensures that the perturbation enhancement is always spatially aligned with the key areas of the model's decision, thereby significantly improving the targeting of adversarial attacks.

\subsection{Objective Functions}
In this study, we aim to effectively prevent the generation of Deepfake while forming perturbed images with high visual quality. To maintain perceptual fidelity during perturbation, we formulate the optimization problem using an $L_1$ distance metric between corresponding pixels, which is defined as:
\begin{equation}
\mathcal{L}_{\mathrm{MAE}} = ||I_{adv}-I_{ori}||_1,
\label{eq:mae}
\end{equation}
where $I_\textrm{ori}$  represents the original image, and $I_\textrm{adv}$ represents the perturbed image.

The $L_2$ norm is adopted as a constraint to regulate the differences between the unperturbed and perturbed images, both in their original and generated forms. The loss function is defined as
\begin{equation}
\mathcal{L}_\textrm{MSE} = -\| I'_\textrm{ori} - I'_\textrm{adv} \|_2^2,
\end{equation}
where $I'_\textrm{ori}$ and $I'_\textrm{adv}$ represent the generated results of $I_\textrm{ori}$ and $I_\textrm{adv}$ after the generative model is applied.

Adversarial Attention Loss function dynamically focuses on critical regions through a difference region mask, maximizing the attention discrepancy in these key areas to optimize perturbation generation. This ensures that the generated fake images exhibit significant differences in the attention regions of the Vision Transformer (ViT)~\cite{dosovitskiy2020image} following
\begin{equation}
    \mathcal{L}_{\text{cam}} = - \log \left( \frac{\sum M_{\text{diff}} \cdot |H'_{\text{ori}} - H'_{\text{adv}}|}{\sum M_{\text{diff}}} \right),
\end{equation}
and
\begin{equation}
\label{M_diff}
    M_{\text{diff}} = \mathbb{I}(|H'_{\text{ori}} - H'_{\text{adv}}| > T),
\end{equation}
where $H'_\textrm{ori}$ and $H'_\textrm{adv}$ are the Grad-CAM heatmaps generated by the ViT network, corresponding to the generated image $I'_\textrm{ori}$ and the perturbed image $I'_\textrm{adv}$, respectively. 
$M_\textrm{diff}$ is a binary mask that highlights regions where the absolute difference between  $H'_\textrm{ori}$ and $H'_\textrm{adv}$ exceeds a threshold $T$.

The total loss is computed as: 

\begin{equation}
\label{total_loss}
\mathcal{L}_{\text{total}} = \lambda_1 \mathcal{L}_{\text{MAE}} + \lambda_2 \mathcal{L}_{\text{MSE}} + \lambda_3 \mathcal{L}_{\text{cam}},
\end{equation}
where $\lambda_1$, $\lambda_2$, and $\lambda_3$ are the weights that determine the inffuence of each loss component.

\begin{table}[htbp]
    \centering
    \caption{Quantitative visual quality evaluation of perturbed images on the CelebA-HQ dataset. Best performance marked in \textbf{bold}.}
    \label{celebA-HQ_shijue}
    \begin{tabular}{@{}l@{\hspace{2em}}c@{\hspace{2em}}c@{\hspace{2em}}c@{}}
    \toprule
    Model         & PSNR↑            & SSIM↑           & LPIPS↓          \\ \midrule
    CMUA~\cite{huang2022cmua}          & 38.6395          & 0.9504          & 0.0333          \\
    Anti-Forgery~\cite{wang2022anti}   & 38.0704          & 0.9529          & 0.0281          \\
    Initiative~\cite{huang2021initiative}    & 35.8042          & 0.9114          & 0.0866          \\
    DF-RAP~\cite{qu2024df}       & 38.8466          & 0.9349          & 0.0511          \\
    
    Ours & \textbf{39.9355} & \textbf{0.9617} & \textbf{0.0251} \\ \bottomrule
    \end{tabular}
\end{table}

\begin{table*}[htbp]
\vspace*{10pt} 
\centering

\caption{Qualitative evaluation of the mean $L_2$ norm distance and DSR between $I'_\textrm{ori}$ and $I'_\textrm{adv}$ on the CelebA-HQ dataset. The best result is marked in \textbf{bold}.}
\label{celebA-HQ_res}

\renewcommand{\arraystretch}{1.2}
\begin{tabularx}{\linewidth}{l *{10}{>{\centering\arraybackslash}X}} 
\toprule
\multirow{2}{*}{Models} & \multicolumn{2}{c}{CMUA~\cite{huang2022cmua}} & \multicolumn{2}{c}{Anti-Forgery~\cite{wang2022anti}} & \multicolumn{2}{c}{Initiative~\cite{huang2021initiative}} & \multicolumn{2}{c}{DF-RAP~\cite{qu2024df}} & \multicolumn{2}{c}{Ours} \\ 
\cmidrule(lr){2-3} \cmidrule(lr){4-5} \cmidrule(lr){6-7} \cmidrule(lr){8-9} \cmidrule(lr){10-11}
                        & $L_2$↑ & DSR↑ & $L_2$↑ & DSR↑ & $L_2$↑ & DSR↑ & $L_2$↑ & DSR↑ & $L_2$↑ & DSR↑ \\ 
\midrule
StarGAN~\cite{choi2018stargan}       & 0.045 & 35.46\% & \textbf{0.484} & \textbf{100\%} & 0.152 & 98.80\% & 0.006 & 0.23\% & 0.340 & \textbf{100\%} \\
AttGAN~\cite{he2019attgan}         & 0.041 & 42.82\% & 0.017 & 0.07\% & 0.010 & 0.0\% & 0.002 & 0.0\% & \textbf{0.046} & \textbf{51.22\%} \\
AGGAN~\cite{tang2019attention}      & 0.041 & 48.01\% & 0.004 & 0.02\% & 0.006 & 0.08\% & 0.003 & 0.02\% & \textbf{0.161} & \textbf{100\%} \\
HiSD~\cite{li2021image}           & 0.052 & 54.77\% & 0.004 & 0.04\% & 0.016 & 0.47\% & 0.007 & 0.10\% & \textbf{0.079} & \textbf{64.79\%} \\
StarGAN-V2~\cite{choi2020stargan} & 0.029 & 14.83\% & 0.040 & 23.23\% & 0.038 & 21.28\% & 0.027 & 12.28\% & \textbf{0.097} & \textbf{93.06\%} \\
\midrule
Average                              & 0.042 & 39.18\% & 0.110 & 24.67\% & 0.044 & 24.13\% & 0.009 & 2.53\% & \textbf{0.145} & \textbf{81.81\%} \\
\bottomrule
\end{tabularx}
\end{table*}

\section{EXPERIMENTS}
\subsection{Implementation Details and Metrics}
\noindent\textbf{Deepfake Models}. 
We validated our framework against leading adversarial models, including StarGAN~\cite{choi2018stargan}, AttGAN~\cite{he2019attgan}, AGGAN~\cite{tang2019attention}, and HiSD~\cite{li2021image} for attribute editing and StarGAN-V2~\cite{choi2020stargan} for expression reenactment. StarGAN and AGGAN were trained for multi-attribute facial editing, covering five key attributes: black hair, blond hair, brown hair, gender, and age. AttGAN supports a broader range of 14 attributes, including skin color, bangs, beard, etc., while HiSD specializes in glasses manipulation.

\noindent\textbf{Datasets}. 
We conducted experiments using the CelebA-HQ~\cite{karras2017progressive} dataset, which consists of 30,000 high-resolution (1024$\times$1024) image samples and features 6,217 unique identities, covering diverse facial attributes. The dataset was partitioned into training, validation, and testing sets following the official split. For cross-dataset evaluation, we utilized the LFW~\cite{huang2012learning} dataset, a widely adopted benchmark for unconstrained face verification and recognition. The LFW dataset contains 5,749 distinct identities, making it suitable for assessing generalization performance. In our experiments, all facial images are resized to 256$\times$256.

\noindent\textbf{Parameters}. 
For the bilateral filter, $\Omega=31$ represents the neighborhood window range, $\sigma_d=75$ is the standard deviation of the spatial Gaussian kernel, and $\sigma_i=15$ is the standard deviation of the range Gaussian kernel. We selected $P=8$ sampling points and a neighborhood radius of $R=1$ to compute the LBP feature map. The threshold $T$ in Eqn.(\ref{M_diff}) is set to 0.3. For the coefficients in Eqn.(\ref{total_loss}), we set
$\lambda_1=1.0$, $\lambda_2=0.04$, and $\lambda_3=0.1$ for total loss.

\noindent\textbf{Evaluation Metrics}. 
We evaluated the visual quality of perturbed images using three metrics, peak signal-to-noise ratio (PSNR) to measure pixel-level fidelity, structural similarity index (SSIM) to assess structural preservation, and learned perceptual image patch similarity (LPIPS) to quantify perceptual differences aligned with human visual perception. To evaluate the effectiveness of the proposed defense method, we computed the visual distance between the original tampered image and perturbation-protected tampered image using the $L_2$ norm. Additionally, we utilized the Defense Success Rate (DSR), which records the percentage of disrupted images exhibiting distortions with $L_2 \geq 0.05$ as a metric to measure the efficacy of the defense.

\subsection{Visual Quality}
Table~\ref{celebA-HQ_shijue} presents a comparative evaluation of five defense methods across three image quality metrics: PSNR, SSIM, and LPIPS. Our method consistently achieves superior performance across all metrics. Specifically, it attains the highest PSNR value of 39.9355, being the only method among the compared approaches to surpass 39 dB. For structural similarity, our method achieves an SSIM score of 0.9617, outperforming Anti-Forgery by 0.0088. In terms of perceptual similarity, we further reduce the LPIPS score by 10.7\% compared to the second-best method, indicating higher visual fidelity and better alignment with human perception.

Fig.~\ref{fig2} visualizes the perturbed images generated by each method and the corresponding Deepfake-forged images. Specifically, in order to intuitively compare the visual quality of the perturbed images and the defense effect of the Deepfake generated images, we showed the original input images and the normal Deepfake results in the first row. It is worth noting that we add a source image in the last column, which is used to provide expression features in StarGAN-V2. Every two remaining columns show the perturbed images generated by different perturbation methods and the defense effects of the corresponding perturbed images after attribute editing or expression reenactment. It can be observed that the perturbed images generated by the Initiative method present an unnatural green effect, the perturbed images from DR-RAP display unnatural artifacts that are visually noticeable, and our method introduces obvious distortions while maintaining excellent visual quality.

\begin{table*}[htbp]
\vspace*{10pt} 
\centering
\caption{Qualitative evaluation of the mean $L_2$ norm distance and DSR between $I'_\textrm{ori}$ and $I'_\textrm{adv}$ on the LFW dataset. The best result is marked in \textbf{bold}.}
\label{LFW_duikang}

\renewcommand{\arraystretch}{1.2}
\begin{tabularx}{\linewidth}{l *{8}{>{\centering\arraybackslash}X}} 
\toprule
\multirow{2}{*}{Models} & \multicolumn{2}{c}{CMUA~\cite{huang2022cmua}} & \multicolumn{2}{c}{Initiative~\cite{huang2021initiative}} & \multicolumn{2}{c}{DF-RAP~\cite{qu2024df}} & \multicolumn{2}{c}{Ours} \\ 
\cmidrule(lr){2-3} \cmidrule(lr){4-5} \cmidrule(lr){6-7} \cmidrule(lr){8-9}
                        & $L_2$↑ & DSR↑ & $L_2$↑ & DSR↑ & $L_2$↑ & DSR↑ & $L_2$↑ & DSR↑ \\ 
\midrule
StarGAN~\cite{choi2018stargan}       & 0.055 & 57.52\% & 0.035 & 29.50\% & 0.004 & 0.01\%  & \textbf{0.177} & \textbf{100\%}    \\
AttGAN~\cite{he2019attgan}         & 0.038 & 12.57\% & 0.004 & 0.02\%  & 0.009 & 0.11\%  & \textbf{0.042} & \textbf{50.57\%}  \\
AGGAN~\cite{tang2019attention}      & 0.048 & 41.50\% & 0.043 & 40.84\% & 0.002 & 0.0\%   & \textbf{0.072} & \textbf{90.75\%}  \\
HiSD~\cite{li2021image}           & 0.045 & 39.09\% & 0.003 & 0.0\%   & 0.001 & 0.0\%   & \textbf{0.069} & \textbf{60.32\%}  \\
StarGAN-V2~\cite{choi2020stargan} & 0.056 & 47.55\% & 0.066 & 61.92\% & 0.030 & 14.69\% & \textbf{0.143} & \textbf{99.72\%}  \\
\midrule
Average                              & 0.048 & 39.65\% & 0.030 & 26.46\% & 0.009 & 2.96\%  & \textbf{0.101} & \textbf{80.27\%}  \\
\bottomrule
\end{tabularx}
\end{table*}

\subsection{Comparison with SOTA}
In Table~\ref{celebA-HQ_res}, we compared the mean $L_2$ norm distance and DSR performance with four popular perturbation methods, namely, CMUA~\cite{huang2022cmua}, Anti-Forgery~\cite{wang2022anti}, Initiative~\cite{huang2021initiative}, and DF-RAP~\cite{qu2024df}.
As demonstrated in Table~\ref{celebA-HQ_res}, our method achieves superior performance compared to other state-of-the-art models across multiple metrics. Specifically, when defending against StarGAN attacks, both Anti-Forgery and Initiative exhibit strong defensive capabilities. Although the $L_2$ distance of our perturbation is slightly lower than Anti-Forgery’s, it still achieves a 100\% defense success rate, matching the best existing perturbation methods. When fighting against AttGAN, our method fails to cause serious distortions in the results, but it produces perceptible manipulation traces in the visualizations. In terms of the average attack effect of the five models, our method surpasses the second-best method by 42.63\% in defense success rate.

We used the attributes of blond hair, beard, young, and glasses to visualize the adversarial effects of StarGAN, AttGAN, AGGAN, and HiSD. Our perturbations induce visible distortions in the generated images across all Deepfake models, disrupting their ability to produce natural-looking results. For StarGAN-V2, it transfers facial expressions from a source image to a target image while preserving the target’s hairstyle and background. In our experimental design, all perturbations are added to the target image, so the generated results of StarGAN-V2 reenact the source expressions on the target images. The results show that none of the four defense methods we compared can resist the manipulation of StarGAN-V2, our approach uniquely compromised its generation process. Although our perturbations do not introduce noticeable artifacts in facial regions, they successfully cause severe distortions in hairstyles and background elements.

\begin{table}[htbp]
    \centering
    \caption{Quantitative visual quality evaluation of perturbed images on the LFW dataset. The best result is marked in \textbf{bold}. }
    \label{LFW_shijue}
    \begin{tabular}{@{}l@{\hspace{2em}}c@{\hspace{2em}}c@{\hspace{2em}}c@{}}
    \toprule
    Model         & PSNR↑            & SSIM↑           & LPIPS↓          \\ \midrule
    CMUA~\cite{huang2022cmua}           & 38.9998          & 0.9497          & 0.0581          \\
    Initiative~\cite{huang2021initiative}     & 36.8727          & 0.8775          & 0.1325          \\
    DF-RAP~\cite{qu2024df}        & 37.1954          & 0.9179          & 0.1241           \\
    Ours & \textbf{39.9101} & \textbf{0.9545} & \textbf{0.0195} \\ \bottomrule
    \end{tabular}
\end{table}

In order to further prove the effectiveness of our method on unseen datasets, we used LFW for cross-dataset verification. Since the LFW dataset does not include attribute annotations for the original images, we omitted the Anti-Forgery approach in our experiments. As shown in Table~\ref{LFW_shijue}, our method effectively preserves the visual quality and texture structure of the original protected image, and can still show the best visual quality on unseen datasets. As shown in Table~\ref{LFW_duikang}, CMUA shows a certain defense ability when fighting against five models, but neither the single defense nor the average defense effects can outperform our method. This result illustrates the generalization ability and robustness of our method, proving that it has practical application potential in the field of perturbation defense.

\subsection{Ablation Study}
We conducted separate experiments using ResNet50 and ViT to generate decision key maps for guiding perturbation generation. This single-model approach allows us to verify the effectiveness of each model's key components in perturbation generation, and examine how the Grad-CAM-guided attention mechanism operates in isolation. We first tested ResNet50 independently, then repeated the experiments using ViT alone.

Following the same evaluation process as the previous experiment, the results of each training are exhibited in Table~\ref{xiaorong}. In terms of visual quality, ResNet50-only maintains similar visual effects to our model, but ViT-only lacks the local attention guidance of ResNet50. The LBP texture feature cannot accurately locate the high-frequency areas that the model is sensitive to. The perturbation enhancement becomes a global uniform injection, resulting in the smooth area also being covered by the perturbation, affecting the visual quality. In terms of adversarial effects, the average defense success rate of ResNet50-only decreased by approximately 8\% due to the lack of loss calculation of the difference area mask. While ResNet50-only achieves marginally higher visual quality scores, the improvement is trivial and visually undetectable. In contrast, our method achieves superior $L_2$ and DSR performance, establishing the optimal balance between fidelity and defense performance.

\begin{table}[htbp]
\centering
\caption{The visual quality and defense effectiveness under different settings.}
\label{xiaorong}
\setlength{\tabcolsep}{3pt} 
\begin{tabularx}{0.95\linewidth}{l *{6}{>{\centering\arraybackslash}X}} 

\toprule
\multirow{2}{*}{Models} & \multicolumn{2}{c}{ResNet50-only} & \multicolumn{2}{c}{ViT-only} & \multicolumn{2}{c}{Ours} \\ 
\cmidrule(lr){2-3} \cmidrule(lr){4-5} \cmidrule(lr){6-7}
                        & L2↑   & DSR↑   & L2↑   & DSR↑   & L2↑   & DSR↑   \\ 
\midrule
StarGAN~\cite{choi2018stargan}                 & 0.318 & 100\%  & 0.405 & 100\%  & 0.340 & 100\%  \\
AttGAN~\cite{he2019attgan}                  & 0.022 & 32.27\% & 0.042 & 50.35\% & 0.046 & 51.22\% \\
AGGAN~\cite{tang2019attention}                   & 0.101 & 98.82\% & 0.017 & 100\%   & 0.161 & 100\%   \\
HiSD~\cite{li2021image}                    & 0.052 & 56.14\% & 0.072 & 60.11\% & 0.079 & 64.79\% \\
StarGAN-V2~\cite{choi2020stargan}             & 0.082 & 81.25\% & 0.090 & 89.34\% & 0.097 & 93.06\% \\ 
\midrule
Average                     & 0.115 & 73.70\% & 0.125 & 79.96\% & 0.145 & 81.81\% \\ 
\midrule
PSNR↑                   & \multicolumn{2}{c}{39.9547} & \multicolumn{2}{c}{37.5541} & \multicolumn{2}{c}{39.9355} \\
SSIM↑                   & \multicolumn{2}{c}{0.9688}  & \multicolumn{2}{c}{0.9577}   & \multicolumn{2}{c}{0.9617}   \\
LPIPS↓                  & \multicolumn{2}{c}{0.0279}  & \multicolumn{2}{c}{0.0446}   & \multicolumn{2}{c}{0.0251}   \\
\bottomrule
\end{tabularx}
\end{table}

\section{CONCLUSIONS}
In this paper, we propose a facial texture-aware perturbation generation framework with dual-branch collaborative optimization, which proactively defends against malicious Deepfake manipulations. After initially extracting texture features using LBP, we identify key image areas via a perturbation enhancement module integrated with Grad-CAM, and guide perturbation generation under local detail enhancement and global semantic constraints. This method achieves a balance between adversarial resistance and visual fidelity. Experiments demonstrate that our framework is effective in disrupting a variety of Deepfake models. The generated forged images exhibit significant distortions, while the original images retain imperceptible perturbations that do not affect human perception, demonstrating the robustness and versatility of our approach across different scenarios.

\bibliographystyle{IEEEtran}
\bibliography{root}

\end{document}